\documentclass{article}

\usepackage[preprint]{corl_2026} 
\usepackage{graphicx}
\usepackage{amsmath}
\usepackage{amssymb}
\usepackage{booktabs, multirow}
\usepackage{subcaption}
\usepackage{pifont}
\usepackage[table]{xcolor} 
\usepackage{pgfplots}
\pgfplotsset{compat=1.14}
\usepackage{algorithm}
\usepackage{algorithmic}
\title{WAM-Nav: Asymmetric Latent World-Action Modeling for Unified Visual Navigation}

\author{
  \textbf{Ning Yang}$^{1,2}$, \textbf{Yan Huang}$^{2,3,4,\ast}$, \textbf{Kaiwen Peng}$^{2,3}$, \textbf{Ziheng He}$^{2,3}$, \\
  \textbf{Kai Wang}$^{2,3}$, \textbf{Cui Miao}$^{5}$, \textbf{Kailin Lyu}$^{2,3}$, \textbf{Guo Li}$^{7}$, \\
  \textbf{Xiaofeng Wang}$^{6,7}$, \textbf{Zheng Zhu}$^{7,\ast}$, \textbf{Jing Liu}$^{5}$, \textbf{Nianfeng Liu}$^{5}$ \\[10pt]
    $^{1}$Nanjing University \quad $^{2}$Institute of Automation, Chinese Academy of Sciences \\
    $^{3}$University of Chinese Academy of Sciences \quad $^{4}$FiveAges \\
    $^{5}$National University of Defense Technology \quad $^{6}$Tsinghua University \quad $^{7}$GigaAI
  }

\begin{document}
\maketitle
\begingroup
\renewcommand\thefootnote{}
\footnotetext{* Corresponding author.}
\addtocounter{footnote}{-1} 
\endgroup
\begin{abstract}
    Visual navigation requires generating smooth and collision-free trajectories under complex geometric and physical constraints. Existing reactive policies that directly map observations to actions lack anticipatory reasoning, limiting their ability to proactively avoid obstacles. While visual imagination offers predictive foresight, conventional modular approaches separate scene prediction from policy learning, often leading to error accumulation and inefficient inference.
    To address these limitations, we propose WAM-Nav, a Latent World-Action Model for embodied visual navigation that jointly learns action generation and latent visual foresight, enabling more robust and foresighted navigation decisions without compromising inference efficiency.
    Specifically, WAM-Nav utilizes a shared Diffusion Transformer for asymmetric joint diffusion to concurrently generate long-horizon actions and short-horizon visual foresight, reducing the inference latency and visual error accumulation inherent in multi-step autoregressive rollouts.
    To further encourage smooth and consistent trajectory generation, we introduce a dual-stream contextual conditioning mechanism that integrates episode-level ego-motion history with sequential visual observations. Combined with a unified goal alignment module that preserves balanced representations across goal types, WAM-Nav naturally supports Image-Goal, Point-Goal, and No-Goal exploration within a single policy.
    Extensive experiments on the challenging \textit{ClutterScenes} and \textit{InternScenes} benchmarks demonstrate strong generalization of WAM-Nav, particularly on Image-Goal and Point-Goal navigation, where it improves success rates by \textbf{15.7\%} and \textbf{3.3\%}, respectively. Real-world deployment further validates effective zero-shot sim-to-real transfer, achieving an average \textbf{85\%} task success rate across diverse indoor and outdoor environments.
\end{abstract}

\keywords{Embodied Visual Navigation, World-Action Model, Diffusion Transformer} 


\section{Introduction}

\begin{figure}[t]
    \centering
    \includegraphics[width=\linewidth]{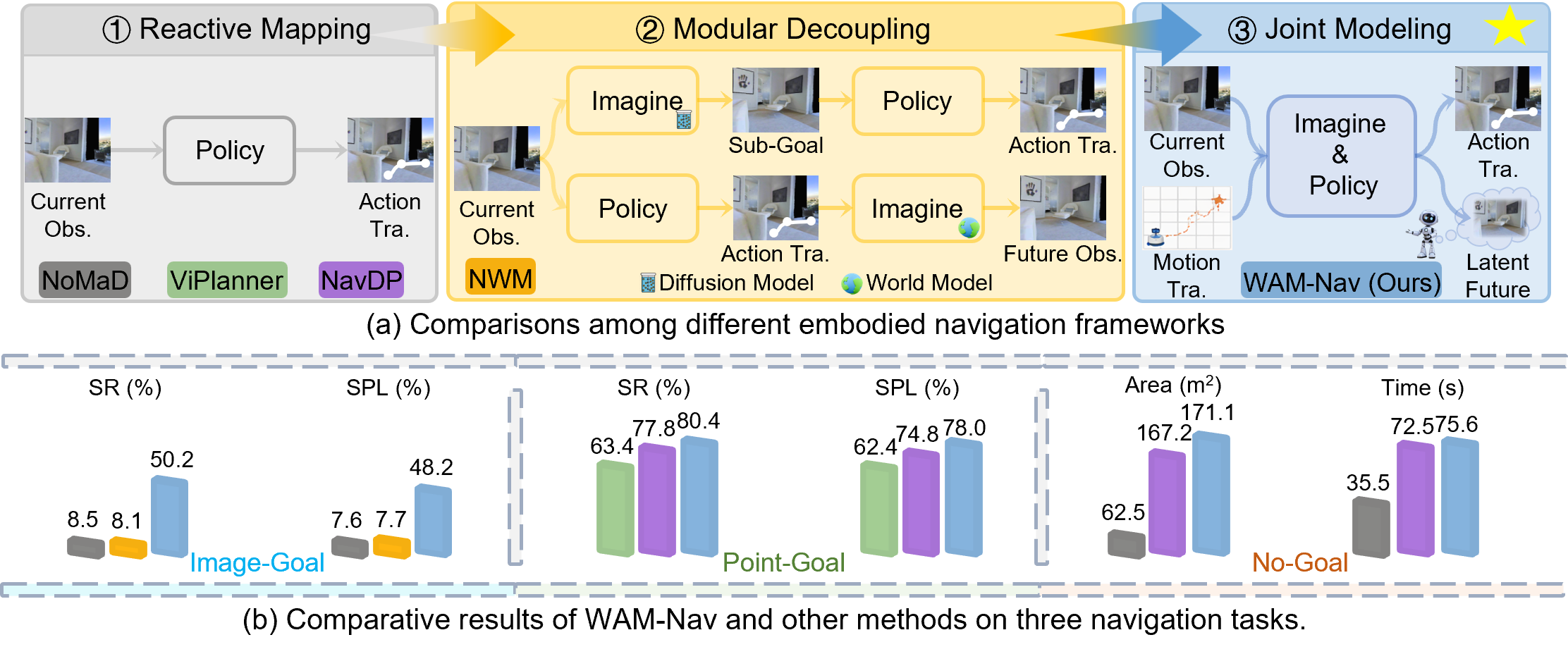}
    \caption{Overview of the \texttt{WAM-Nav} paradigm and performance. (a) Methodological comparison: Unlike traditional purely reactive or decoupled modular pipelines, \texttt{WAM-Nav} jointly models action generation and latent visual foresight within a unified framework. (b) Quantitative performance: Our method achieves competitive results against established baselines across diverse evaluation settings.}
    \label{fig:intro}
    \vspace{-5mm}
\end{figure}

Embodied visual navigation \citep{wei2026navol, yang2025transformer, yang2025longterm, zhang2025navfom} aims to guide agents through complex, unseen physical environments by generating smooth, collision-free trajectories based on visual observations \citep{zhu2021survey}. Over the past few years, the dominant navigation paradigm has gradually transitioned from traditional decoupled mapping-and-planning pipelines \citep{campos2021orb, labbe2019rtabmap, chaplot2020learning} to learning-based intelligent methods. As shown in Figure \ref{fig:intro} (a), recent end-to-end reactive approaches \cite{shah2023gnm, shah2023vint, sridhar2024nomad, peng2025logoplanner, cai2025navdp} directly map visual observations to executable actions. 
Moreover, to endow agents with predictive reasoning capabilities, recent research has actively explored visual imagination and world-model-based navigation strategies. These modular approaches generally fall into two distinct paradigms. The first paradigm prioritizes generating future visual subgoals, which are subsequently utilized to infer control actions via inverse dynamics models \cite{qin2025navigatediff, ni2024generatesubgoal}. The second paradigm focuses on trajectory evaluation, where candidate paths are first generated and then scored by a world model that imagines future visual states\cite{bar2025nwm,zhang2026raenwm, dong2025uniwm}. 

Although existing approaches have improved robotic navigation to varying degrees, fundamental limitations remain across current paradigms. Reactive end-to-end mapping methods, which directly map observations to actions, rely heavily on instantaneous perception and therefore lack predictive reasoning and local dynamics understanding, making them prone to local optima and collisions in cluttered environments. Modular decoupled approaches provide explicit foresight, but their separate training of action decision-making and future imagination introduces substantial computational latency and compounding errors, often resulting in delayed responses and poor trajectory consistency. While existing World-Action Models have shown promising joint visual-action modeling capabilities in robot manipulation \citep{ye2026dreamzero, bi2025motus, antgroup2026lingbotva}, their autoregressive generation paradigm remains challenging for navigation due to limited real-time capability and accumulated prediction errors under large viewpoint changes. Furthermore, most existing navigation methods are designed for a single target specification \citep{shah2023gnm,shah2023vint,sridhar2024nomad,peng2025logoplanner,bar2025nwm,qin2025navigatediff,dong2025uniwm}, requiring redesign, new data collection, and retraining when adapting to new tasks. Although NavDP \citep{cai2025navdp} supports multiple goal-oriented navigation tasks, its unimodal alignment design still leads to imbalanced performance across task types.

Motivated by these advances, we propose \texttt{WAM-Nav}, a Latent World-Action Model for embodied visual navigation that jointly learns action generation and latent visual foresight, enabling more robust and foresighted navigation decisions without compromising inference efficiency, as illustrated in Figure \ref{fig:intro}. Instead of relying on sequential decoupled modules, \texttt{WAM-Nav} utilizes a shared Diffusion Transformer (DiT) for asymmetric joint diffusion to concurrently generate long-horizon actions and short-horizon visual foresight. This joint modeling reduces the inference latency and visual error accumulation inherent in multi-step autoregressive rollouts, while ensuring a deep coupling between visual features and control dynamics to empower the policy to anticipate environmental variations and proactively steer clear of obstacles. To further guide generation and preserve trajectory consistency, we introduce Dual-Stream Contextual Conditioning (DSCC), which integrates episode-level ego-motion history with sequential visual observations. Combined with a unified goal alignment module that preserves balanced representations across goal types, \texttt{WAM-Nav} naturally supports Image-Goal, Point-Goal, and No-Goal exploration within a single policy without complex architecture switching.
We conduct extensive evaluations on cluttered and physically realistic benchmarks, including the \textit{ClutterScenes} and \textit{InternScenes}. Experimental results demonstrate that \texttt{WAM-Nav} achieves strong zero-shot generalization, particularly on Image-Goal and Point-Goal navigation where it improves success rates by \textbf{15.7\%} and \textbf{3.3\%}, respectively. Furthermore, deployment on physical robot platforms validates its capability for sim-to-real transfer, achieving an average \textbf{85\%} task success rate across diverse indoor and outdoor environments.

\section{Related Work}

\subsection{Learning-based Embodied Visual Navigation}

In contrast to traditional mapping-and-planning methods, the end-to-end learning framework has emerged as a dominant paradigm for embodied visual navigation \citep{zhu2021survey, zhu2017target, savva2019habitat, wijmans2020ddppo, zeng2025navidiffusor}. Recent foundation models have significantly advanced this field: GNM \citep{shah2023gnm} demonstrated cross-embodiment generalization using a lightweight CNN policy \citep{krizhevsky2012imagenet}, while ViNT \citep{shah2023vint} extended navigation horizons via Transformer-based sub-goal planning. Subsequently, NoMaD \citep{sridhar2024nomad} introduced diffusion policies \citep{janner2023policy} to model multi-modal action distributions, and NavDP \citep{cai2025navdp} incorporated an actor-critic mechanism to score and select the safest generated trajectories. Despite these advances, such reactive policies directly map local observations to actions without explicitly modeling their future perceptual consequences, rendering them susceptible to collisions in cluttered environments \citep{qin2025navigatediff, shen2026efficient}. In contrast, \texttt{WAM-Nav} addresses this limitation by integrating predictive visual foresight directly into the policy, enabling proactive obstacle avoidance and safer navigation decision-making.

\subsection{World Models and Visual Imagination for Navigation}

Integrating visual imagination facilitates foresight-driven navigation \citep{zhang2026sparsevideonav, nie2025wmnav}, and early systems typically adopt modular architectures to evaluate candidate motions. For instance, Pathdreamer \citep{koh2021pathdreamer} and NWM \citep{bar2025nwm} rely on decoupled visual prediction and trajectory planning, synthesizing future observations via a world model built atop an existing navigation policy. Similarly, NavigateDiff \citep{qin2025navigatediff} employs diffusion models \citep{ho2020denoising} to imagine future visual sub-goals, which are subsequently executed by a separate inverse dynamics policy. These approaches treat visual imagination and trajectory generation as sequential and decoupled processes. This separation frequently introduces state-action misalignment and compounding control errors. 
In response, \texttt{WAM-Nav} jointly models action generation and visual foresight, where future visual dynamics are predicted in a latent space. Rather than decoupling prediction from decision-making, it learns both jointly to promote temporal consistency and physical feasibility with minimal inference latency.

\section{Methodology}
\label{sec:method}
\begin{figure}[t]
    \centering
    \includegraphics[width=\linewidth]{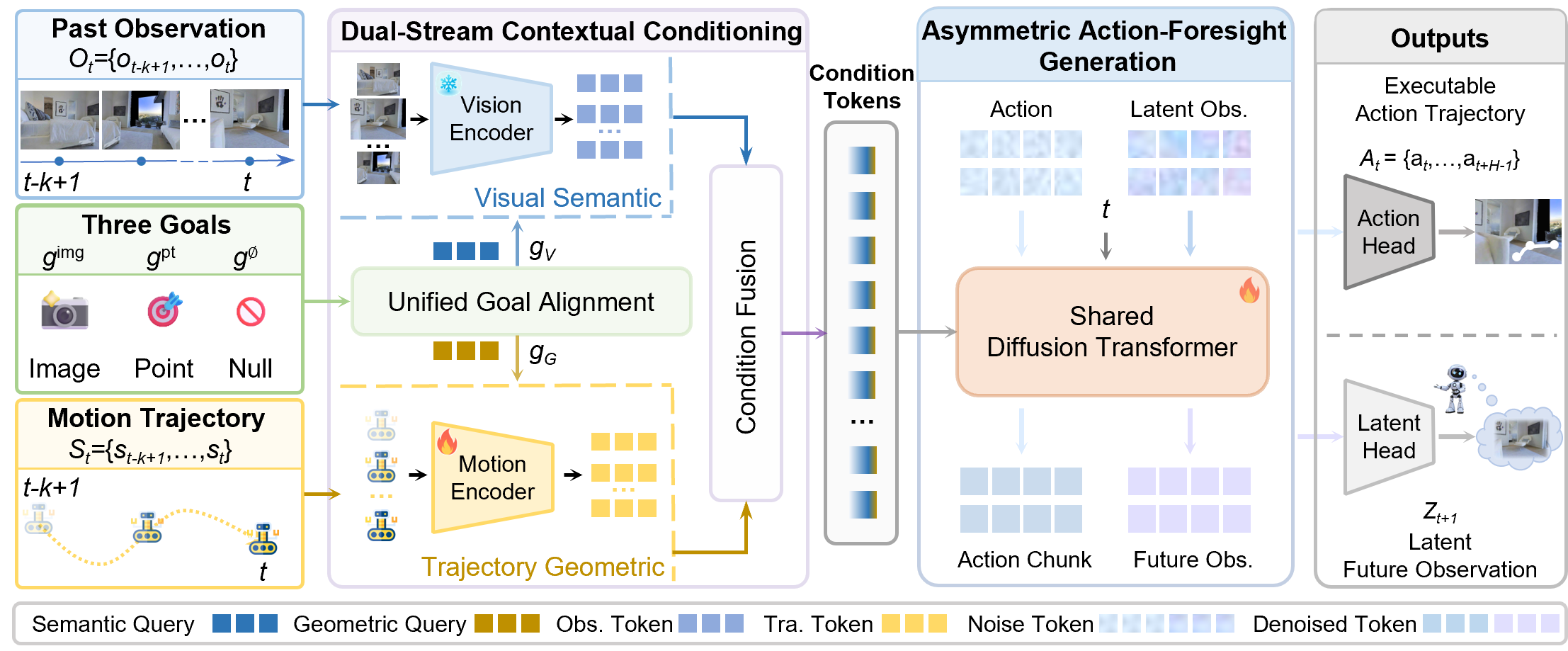}
    \caption{Architecture overview of the \texttt{WAM-Nav} framework. Heterogeneous navigation goals are explicitly routed into visual-semantic ($g_V$) and trajectory geometric ($g_G$) queries. These queries contextually modulate historical RGB-D sequences and relative ego-motion trajectories, synthesizing a compact conditioning context $C$. Conditioned on $C$, a shared DiT performs asymmetric joint generation of future control actions and latent visual foresight.}
    \label{fig:pipeline}
\end{figure}

As illustrated in Figure~\ref{fig:pipeline}, \texttt{WAM-Nav} consists of three key components: (1) \textit{Unified Goal Alignment}, (2) \textit{Dual-Stream Contextual Conditioning}, and (3) \textit{Asymmetric Action-Foresight Generation}. Specifically, to enable balanced performance across diverse navigation tasks, we first project heterogeneous goal specifications into a unified alignment space. We then construct a dual-stream spatiotemporal context by jointly encoding the robot's sequential visual observations and episode-level ego-motion history. Both streams are explicitly modulated by goal-relevant information and aggregated into a compact conditioning context $C$. Conditioned on $C$, a shared DiT jointly generates future action trajectories and latent visual foresight through asymmetric denoising.

\subsection{Unified Goal Alignment}

To enable balanced performance across Image-Goal, Point-Goal, and No-Goal exploration within a single policy, \texttt{WAM-Nav} introduces a unified goal alignment mechanism. Unlike prior navigation approaches~\citep{cai2025navdp, wang2025internvla} that reformulate all tasks as point-goal navigation, our design preserves modality-specific goal information to maintain task-specific expressiveness across different navigation settings. Given a goal input $g$, \texttt{WAM-Nav} encodes it into two complementary goal embeddings: a visual semantic query $g_V \in \mathbb{R}^D$ for visual memory retrieval, and a geometric query $g_G \in \mathbb{R}^D$ for trajectory-level directional guidance.

Formally, the goal input $g$ is first transformed into a base embedding $e_g$ via a modality-specific feature extractor $E_{\phi}(\cdot)$, and then projected into the two functional token spaces:
\begin{equation}
g_V = \psi_V(e_g), \quad g_G = \psi_G(e_g), \quad \text{with} \quad e_g = E_{\phi}(g)
\end{equation}
where $E_{\phi}(\cdot)$ is instantiated according to the goal modality, including a Vision Transformer trained from scratch for image goals, sinusoidal positional encoding for relative coordinates, and a masked zero-state representation for goal-free exploration. The projection operators $\psi_V(\cdot)$ and $\psi_G(\cdot)$ are learnable linear mappings that map the base embedding into the visual semantic and geometric spaces, respectively.

This alignment design preserves modality-specific goal information while providing a unified interface across heterogeneous navigation tasks, enabling balanced performance across different goal settings and producing goal-conditioned context for the subsequent Section~\ref{sec:dscc}, where it serves as conditioning for the joint modeling of action and visual foresight.

\subsection{Dual-Stream Contextual Conditioning}
\label{sec:dscc}

While visual history provides spatial cues for obstacle avoidance, relying solely on visual conditioning often leads to kinematically inconsistent or jittery trajectories due to the lack of explicit momentum constraints. To address this, \texttt{WAM-Nav} introduces Dual-Stream Contextual Conditioning (DSCC) over a historical sliding window from $t-k+1$ to $t$. We dynamically construct the conditioning context $C$ by fusing a goal-modulated visual memory stream (capturing local spatial geometry) with a trajectory-aware kinematic history stream (capturing physical momentum). Fusing these streams provides the shared DiT with a spatiotemporal conditioning context that ensures the generated action trajectories are both geometrically collision-free and kinematically smooth.

\textbf{Goal-Modulated Visual Memory.} 
Historical RGB observations $\mathcal{O}_t$ are processed by DINO-v2 \citep{oquab2023dinov2} into a memory tensor $V \in \mathbb{R}^{B \times (k \cdot P) \times D}$, where $B$ represents the batch size, $k$ is the historical window length, and $P$ is the patch count per frame. To selectively reinforce spatial tokens relevant to the target, the visual query $g_V \in \mathbb{R}^{B \times D}$ computes independent relevance scores for each patch token via a scaled dot-product:
\begin{equation}
    \alpha = \sigma \left( \frac{g_V V^\top}{\sqrt{D}} \right) \in \mathbb{R}^{B \times (k \cdot P) \times 1}
\end{equation}
where $\sigma$ is the sigmoid function. The visual memory is then residually updated:
\begin{equation}
    \tilde{V} = V + \alpha \odot V
\end{equation}

\textbf{Trajectory-Aware Motion History.} 
To incorporate the agent's kinematic momentum and ensure robust zero-shot generalization, the motion trajectory stream processes the executed pose sequence $\mathcal{S}_t$. Absolute poses are converted into coordinate-invariant relative displacements and heading changes in the current egocentric frame, yielding a relative kinematics sequence $\tilde{\mathcal{S}}_t = \{(\Delta x_i, \Delta y_i, \Delta \theta_i)\}_{i=t-k+1}^{t}$ (detailed in Appendix~\ref{appendix:ego_history_conversion}). 

This relative motion sequence $\tilde{\mathcal{S}}_t$ is embedded and processed by a causal Transformer encoder $H \in \mathbb{R}^{B \times k \times D}$ to capture temporal kinematic profiles. To align past momentum with current target directionality, we query $H$ with the geometric goal embedding $g_G \in \mathbb{R}^{B \times D}$ via cross-attention to extract a consolidated kinematic vector $o_{\mathrm{kin}}$:
\begin{equation}
    o_{\mathrm{kin}} = \mathrm{CrossAttn}(g_G, H, H) \in \mathbb{R}^{B \times D}
\end{equation}
The vector $o_{\mathrm{kin}}$ mathematically summarizes the agent's historical motion continuity relative to its target directionality.

\textbf{Condition Fusion via Cross-Attention.} 
To structurally enforce kinematic momentum to guide spatial visual retrieval, the visual and trajectory streams are integrated via a multi-layer Transformer Decoder to formulate the unified conditioning context $C \in \mathbb{R}^{B \times N_c \times D}$. Specifically, the extracted kinematic token $o_{\mathrm{kin}}$ is projected to bias a set of learnable query embeddings $Q_c \in \mathbb{R}^{N_c \times D}$:
\begin{equation}
    C = \mathrm{TransformerDecoder} \left( Q_c + \phi(o_{\mathrm{kin}}), \tilde{V}, \tilde{V} \right)
\end{equation}
where $\phi(\cdot)$ denotes a linear projection. Architecturally, this fusion mechanism utilizes the agent's kinematic momentum ($o_{\mathrm{kin}}$) to actively query the goal-modulated visual spatial context ($\tilde{V}$). The resulting conditioning context $C$ thus explicitly constrains the subsequent latent generation, enforcing both physical execution smoothness and geometric safety.

\subsection{Asymmetric Action-Foresight Generation}

Conditioned on the unified conditioning context $C$, \texttt{WAM-Nav} jointly models the future action trajectory $\mathbf{A}_t = \{a_t, \dots, a_{t+H_{\mathrm{act}}-1}\}$ over a long planning horizon $H_{\mathrm{act}}$ and its corresponding latent visual foresight $\mathbf{Z}_{t+1:t+H_{\mathrm{vis}}} = \{z_{t+1}, \dots, z_{t+H_{\mathrm{vis}}}\}$ over a short prediction horizon $H_{\mathrm{vis}}$ ($H_{\mathrm{vis}} \le H_{\mathrm{act}}$) within a shared representation space. The future visual state $z_i = \mathcal{E}(o_i)$ is compressed via a pre-trained Stable Diffusion VAE \citep{stabilityai2022sd} into a compact grid of $N$ latent patches. 

To optimize this generative process, we employ the flow-matching \citep{streamingflow2025} training strategy to learn the joint conditional distribution $p_\theta(\mathbf{A}, \mathbf{Z} \mid C)$ via asymmetric joint diffusion. This strategy formulates the continuous probability paths as straight lines that map standard Gaussian priors $(\mathbf{A}_0, \mathbf{Z}_0) \sim \mathcal{N}(\mathbf{0}, \mathbf{I})$ to the empirical data manifold. At any flow time $\tau \in [0, 1]$, the interpolated states are defined as:
\begin{equation}
    \mathbf{A}_{\tau} = (1-\tau)\mathbf{A}_0 + \tau \mathbf{A}_1, \quad
    \mathbf{Z}_{\tau} = (1-\tau)\mathbf{Z}_0 + \tau \mathbf{Z}_1
\end{equation}
where the corresponding target velocity fields are given by $u_A = \mathbf{A}_1 - \mathbf{A}_0$ and $u_Z = \mathbf{Z}_1 - \mathbf{Z}_0$.

The architectural backbone for this asymmetric generative progress is a shared DiT. The noised sequences $\mathbf{A}_{\tau}$ and $\mathbf{Z}_{\tau}$ are tokenized, concatenated, and processed through a multi-layer DiT. Crucially, within each DiT block, these heterogeneous tokens undergo a shared self-attention, enabling the dynamic exchange of spatiotemporal constraints between the physical action paths and the latent visual representations at every layer. Both streams cross-attend to the conditioning context $C$, with the timestep $\tau$ modulated via adaptive LayerNorm (adaLN) layers to regress the joint velocity fields:
\begin{equation}
    \hat{u}_A, \hat{u}_Z = f_{\theta}(\mathbf{A}_{\tau}, \mathbf{Z}_{\tau}, \tau, C).
\end{equation}
This shared parameterization makes latent foresight act as a perception-grounded constraint on action generation, penalizing action--scene inconsistency through the visual velocity-matching loss. Unlike manipulation-oriented WAMs~\citep{ye2026dreamzero, bi2025motus}, where future visual changes are often local and object-centric, navigation involves large egocentric viewpoint changes; long autoregressive visual rollouts would introduce both inference latency and accumulated visual errors that may misguide action generation. \texttt{WAM-Nav} therefore adopts an asymmetric design: long-horizon actions preserve trajectory continuity, while short-horizon latent foresight provides reliable near-future geometric constraints without overextending visual prediction.

\subsection{Model Training and Inference}

We train \texttt{WAM-Nav} end-to-end by minimizing a joint objective consisting of flow-matching velocity regression and multi-modal goal alignment:
\begin{equation}
    \label{eq:l_total}
    \mathcal{L}_{\mathrm{total}} = \mathbb{E}_{\tau, \mathbf{A}_0, \mathbf{Z}_0} \left[ \|\hat{u}_A - u_A\|_2^2 + \lambda_{\mathrm{img}}\|\hat{u}_Z - u_Z\|_2^2 \right] + \lambda_{\mathrm{align}}\mathcal{L}_{\mathrm{align}}
\end{equation}
where $\tau \sim \mathcal{U}(0,1)$, and $\lambda_{\mathrm{img}}, \lambda_{\mathrm{align}}$ are scalar weighting coefficients. The term $\mathcal{L}_{\mathrm{align}}$ represents a symmetric contrastive InfoNCE loss \citep{oord2018representation}. By maximizing the mutual information between cross-space projections of equivalent objectives, this alignment constraint ensures consistency across diverse goal modalities, facilitating the extraction of unified directional and visual cues regardless of the original input format.

During online deployment, \texttt{WAM-Nav} follows a receding-horizon control loop. Consistent with \textsc{NavDP}~\citep{cai2025navdp}, at each step, conditioned on the current conditioning context $C$, the policy samples 16 candidate trajectories. Following prior generative navigation policies~\citep{sridhar2024nomad}, the first sampled trajectory is selected for execution.

\section{Experimental Results}
\label{sec:result}
We conduct extensive zero-shot deployments of \texttt{WAM-Nav} to evaluate its effectiveness, focusing on five research questions: 
\textbf{Q1:} Does \texttt{WAM-Nav} achieve superior zero-shot generalization while maintaining balanced performance across diverse navigation tasks?  
\textbf{Q2:} What mechanisms enable \texttt{WAM-Nav} to navigate highly cluttered environments more effectively than existing baselines?  
\textbf{Q3:} How computationally efficient is \texttt{WAM-Nav}, and can it meet the real-time requirements of online navigation?
\textbf{Q4:} How do the individual components contribute to the overall performance improvement?  
\noindent\textbf{Q5:} Can \texttt{WAM-Nav} transfer to physical platforms for zero-shot real-world navigation?
\subsection{Simulation Experiments}

\noindent\textbf{Training Datasets and Setup.}
Consistent with prior works \citep{peng2025logoplanner,cai2025navdp}, \texttt{WAM-Nav} is trained on the large-scale visual navigation dataset VLN-N1 \citep{wang2025internvla, interndata2025}. VLN-N1 is built upon six categories of 3D scene assets, including Replica~\citep{Straub2019Replica}, Matterport3D~\citep{Chang2017Matterport3D}, Gibson~\citep{Xia2018Gibson}, 3D-FRONT~\citep{fu20203dfront}, HSSD~\citep{Khanna2023HSSD}, and HM3D~\citep{Ramakrishnan2021HM3D}. Leveraging an extensive domain randomization pipeline, the dataset provides over 400 hours of collision-free and smooth first-person navigation trajectories collected in simulation, comprising more than 200K trajectories in total.
\texttt{WAM-Nav} is trained with a learning rate of \(1.5\times10^{-4}\) and the total training cost is approximately \(8\times120\) GPU hours. Additional implementation details are provided in Appendix~\ref{appendix:implementation_details}.

\noindent\textbf{Zero-Shot Evaluation Platform.}
We evaluate \texttt{WAM-Nav} and all baselines in a zero-shot setting on a simulation benchmark built upon IsaacSim, using the wheeled robot ClearPath Dingo as the navigator. All methods are directly evaluated with their original pretrained weights, without additional finetuning. The benchmark includes two categories of environments: \textit{ClutterScenes} and \textit{InternScenes} \citep{cai2025navdp, wang2025internvla, wang2024grutopia}.
The \textit{ClutterScenes} contain 10 \textit{easy} scenes and 10 \textit{hard} scenes. Both are constructed from randomly generated layouts with cluttered obstacles, while the \textit{hard} scenes feature denser obstacle distributions. The \textit{InternScenes} include 20 \textit{home} scenes and 20 \textit{commercial} scenes. \textit{Home} scenes are characterized by narrow passages and cluttered layouts, whereas \textit{commercial} scenes cover representative indoor categories such as \textit{supermarkets} and \textit{restaurants}. For each scene, we randomly sample 100 navigation episodes, yielding a total of 6,000 evaluation episodes. 

\noindent\textbf{Metrics.}
For Image-Goal and Point-Goal navigation, we evaluate \textbf{Success Rate (SR)} and \textbf{Success weighted by Path Length (SPL)} to measure task completion and navigation efficiency. For No-Goal exploration, we report episode \textbf{Time} and explored \textbf{Area} to evaluate exploration capability. Episode \textbf{Time} is capped at 120 seconds, and episodes that become locally stuck are terminated early. Higher values indicate better performance for all metrics.

\noindent\textbf{Compared Methods.}

To comprehensively evaluate the effectiveness of \texttt{WAM-Nav}, we benchmark our model against a diverse set of strong baselines tailored to different navigation tasks. Specifically, for \textbf{Image-Goal} navigation, we compare \texttt{WAM-Nav} against GNM~\citep{shah2023gnm}, ViNT~\citep{shah2023vint}, NoMaD~\citep{sridhar2024nomad}, NWM~\citep{bar2025nwm}, and NavDP~\citep{cai2025navdp}. For \textbf{Point-Goal} navigation, the baselines include DD-PPO~\citep{wijmans2020ddppo}, iPlanner~\citep{yang2023iplanner}, ViPlanner~\citep{roth2024viplanner}, and NavDP~\citep{cai2025navdp}. Finally, for \textbf{No-Goal} exploration, we compare against GNM~\citep{shah2023gnm}, ViNT~\citep{shah2023vint}, NoMaD~\citep{sridhar2024nomad}, and NavDP~\citep{cai2025navdp} (detailed configurations are provided in Appendix~\ref{appendix:baseline_details}).

\noindent\textbf{Result Analysis.}
For \textbf{Q1}, Table~\ref{tab:all_tasks_fixed_vertical} (full results in Appendix~\ref{appendix:main_results}) summarizes the zero-shot evaluation results across Image-Goal, Point-Goal, and No-Goal exploration. \texttt{WAM-Nav} achieves the best average performance across the three task settings, reaching 50.2\% SR / 48.2\% SPL on Image-Goal, 80.4\% SR / 78.0\% SPL on Point-Goal, and 171.1 m$^2$ explored area in No-Goal exploration. The consistent gains across both \textit{ClutterScenes} and \textit{InternScenes} suggest that jointly learning action generation with latent visual foresight improves zero-shot robustness under diverse obstacle layouts and semantic scene complexity. Meanwhile, the balanced performance across goal types supports the effectiveness of unified goal alignment for a single multi-task navigation policy.

\begin{figure}[t]
    \centering
    \includegraphics[width=\linewidth]{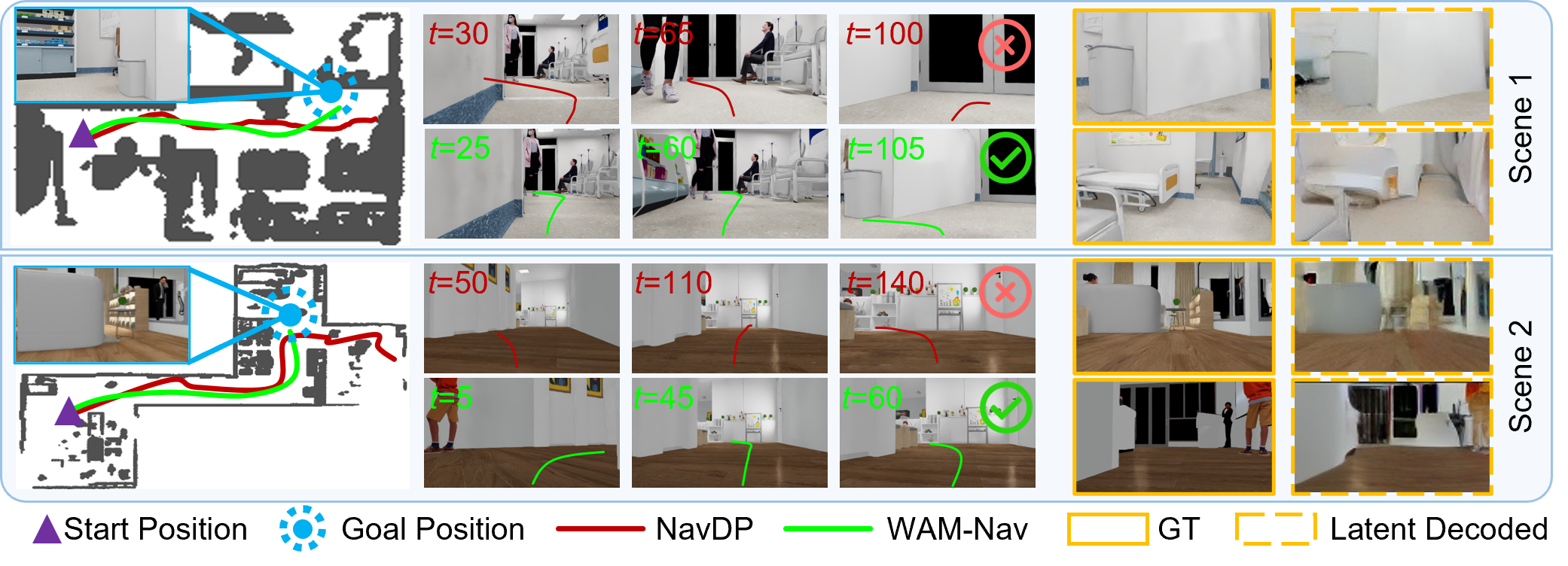}
    \caption{\textbf{Qualitative comparison on image-goal navigation.} As shown in the 2D top-down trajectories and egocentric path projections, compared with NavDP, \texttt{WAM-Nav} exhibits smoother, more consistent trajectories and better real-time obstacle avoidance predictions. Despite generating future states entirely in a compressed latent space, its decoded visual foresights remain highly faithful to ground-truth (GT) observations.}
    \label{fig:result}
    \vspace{-5mm}
\end{figure}

For \textbf{Q2}, Figure~\ref{fig:result} qualitatively compares NavDP and \texttt{WAM-Nav} on the Image-Goal task. NavDP often reacts only after approaching obstacles, causing abrupt trajectory changes in cluttered areas. In contrast, \texttt{WAM-Nav} uses short-horizon latent foresight to anticipate near-future geometric constraints while maintaining long-horizon action continuity, leading to smoother and more proactive obstacle avoidance. The decoded foresight remains consistent with ground-truth observations, indicating that latent foresight provides reliable guidance without requiring long autoregressive visual rollouts.

For \textbf{Q3}, Table~\ref{tab:efficiency_comparison} shows that \texttt{WAM-Nav} remains efficient for online navigation despite incorporating latent visual foresight. Although it has a larger total parameter count due to frozen visual priors, its trainable parameters are comparable to NavDP while requiring lower per-decision computation (0.7 vs. 1.3 TFLOPs). With an inference latency of 0.26\,s, \texttt{WAM-Nav} introduces only modest overhead over NavDP, while avoiding the costly multi-candidate visual rollouts of NWM (1.43\,s, 8.3 TFLOPs), validating the efficiency of asymmetric short-horizon foresight for real-time navigation.

For \textbf{Q4}, Table~\ref{tab:ablation_study_flattened_onedecimal_final} (Appendix~\ref{appendix:main_ablation}) evaluates the roles of trajectory conditioning and latent foresight. Latent prediction alone improves SR from 42.1\% to 45.7\%, showing that foresight helps the policy anticipate local obstacle constraints. Motion trajectories alone improve performance in \textit{ClutterScenes} but degrade in more semantically complex \textit{InternScenes}, indicating that kinematic history improves smoothness but cannot replace perceptual foresight. Combining both yields the best result (50.2\% SR / 48.2\% SPL), supporting the DSCC design: ego-motion history stabilizes trajectory generation, while latent foresight provides the geometric constraints needed for safe navigation.

\begin{table*}[t]
\centering
\caption{Computational efficiency comparison across navigation methods.}
\label{tab:efficiency_comparison}

\small
\renewcommand{\arraystretch}{0.8}

\begin{tabular*}{\textwidth}{@{\extracolsep{\fill}}lcccc}
\toprule

Method
& Inference Latency (s) $\downarrow$
& Total Params
& Trainable Params
& TFLOPs $\downarrow$ \\

\midrule

NWM & 1.43 & 154.4M & 68.3M & 8.3 \\

NavDP
& \textbf{0.16}
& 136.4M
& 136.4M
& 1.3 \\

Ours
& 0.26
& 234.9M
& 129.2M
& \textbf{0.7} \\

\bottomrule
\end{tabular*}

\end{table*}

\begin{figure}[!t]
    \centering
    \includegraphics[width=\linewidth]{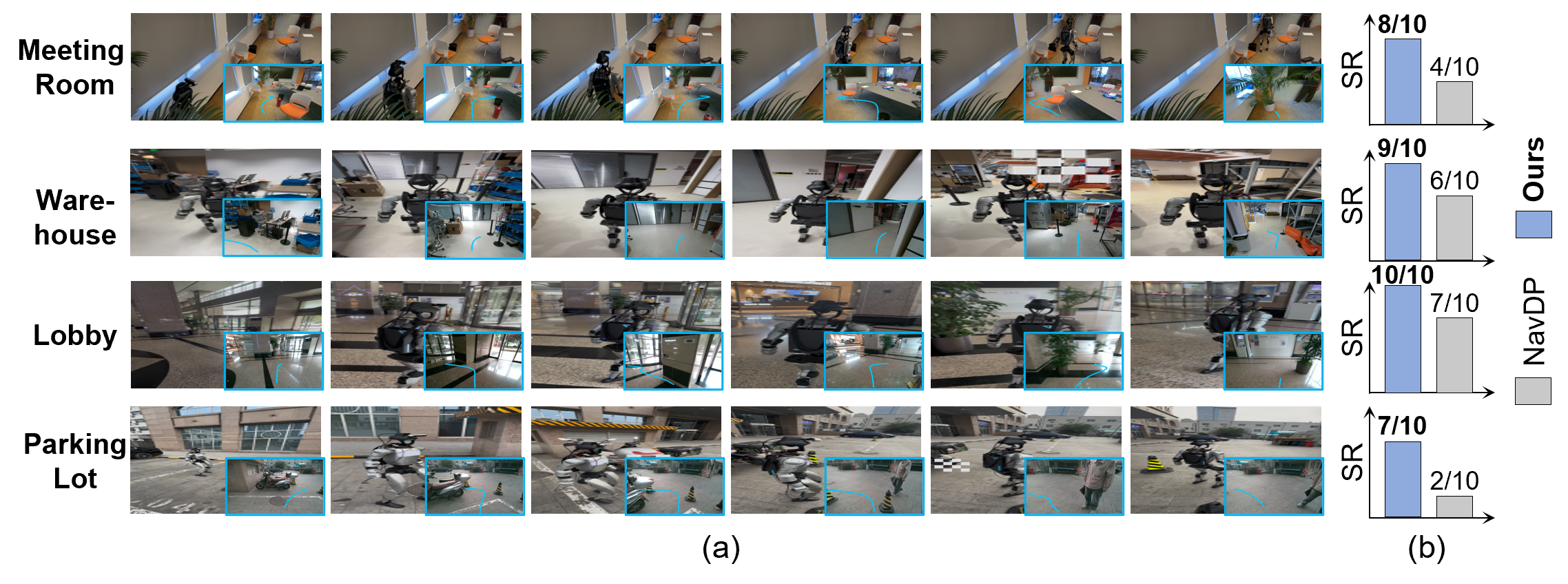}
    \caption{\textbf{Zero-shot deployment of \texttt{WAM-Nav} in real-world environments.} (a) Visualization of \texttt{WAM-Nav} navigating across four representative indoor and outdoor scenes, with the first-person trajectory planning visualization shown in the bottom-right corner of each image. (b) Quantitative comparison of \texttt{WAM-Nav} and NavDP in each real-world scenario.}
    \label{fig:realworld}
    \vspace{-1mm}
\end{figure}

\subsection{Deployment on Real World}


For \textbf{Q5}, we deploy \texttt{WAM-Nav} in a zero-shot manner on a real-world Unitree G1 humanoid robot equipped with an Intel RealSense D455 camera. Evaluation is conducted across four indoor and outdoor environments: a \textit{meeting room}, \textit{warehouse}, \textit{lobby}, and \textit{parking lot}, with 10 trials per scene. As shown in Fig.~\ref{fig:realworld}(a), \texttt{WAM-Nav} consistently predicts feasible trajectories under diverse layouts and lighting conditions, achieving an average 85\% success rate. Compared with NavDP, the higher success rate indicates that latent foresight and motion-aware conditioning improve obstacle avoidance and trajectory stability beyond simulation, supporting effective zero-shot sim-to-real transfer.


\section{Conclusion}
\label{sec:conclusion}

This paper presents \texttt{WAM-Nav}, a Latent World-Action Model for embodied visual navigation that unifies action generation and latent visual foresight within a shared generative framework. By using asymmetric action-foresight generation, \texttt{WAM-Nav} preserves long-horizon trajectory continuity while avoiding the latency and error accumulation of long visual rollouts. Its dual-stream conditioning further integrates sequential visual observations with episode-level ego-motion history, guiding the policy toward trajectories that are both geometrically safe and kinematically smooth. Together with unified goal alignment, the same policy supports Image-Goal, Point-Goal, and No-Goal exploration. Extensive zero-shot experiments on \textit{ClutterScenes} and \textit{InternScenes}, together with real-world deployment on a humanoid robot, demonstrate strong generalization, effective sim-to-real transfer, and practical potential for robust embodied navigation.

\textbf{Limitations.}
During real-world deployment of \texttt{WAM-Nav}, we identify two primary failure modes. First, the limited camera height and field of view restrict perception of near-field obstacles, which may lead to inaccurate trajectory prediction or delayed avoidance. Second, the current policy does not explicitly model robot embodiment, causing mismatches between camera-level obstacle avoidance and full-body traversability. These limitations suggest two promising future directions: adaptive viewpoint control for dynamic perception adjustment, and embodiment-aware training with diverse robot morphologies to improve navigation robustness.

\clearpage


\bibliography{main}  

\newpage
\appendix
\section{Overview}

The supplementary material is organized as follows:
\begin{itemize}
    \item Section B provides the formal problem definition.
    \item Section C provides additional details of the proposed method.
    \item Section D presents the model configurations and training details of \texttt{WAM-Nav}.
    \item Section E introduces the details of the baseline methods.
    \item Section F includes additional experimental results and ablation studies.
    \item Section G describes the hardware setup and deployment details for real-world experiments.
    \item Section H provides the visualization of a failure case.
\end{itemize}

\section{Problem Definition}
\label{sec:problem}

We consider mapless visual navigation under three goal modalities: Image-Goal ($g^{\mathrm{img}}$), Point-Goal ($g^{\mathrm{pt}}$), and No-Goal exploration ($g^{\varnothing}$). At each time step $t$ in a partially observable environment, the agent receives a visual history $\mathcal{O}_t=\{o_{t-k+1}, \ldots, o_t\}$, an action-accumulated ego-motion history $\mathcal{S}_t=\{s_{t-k+1}, \ldots, s_t\}$, and a goal condition $g \in \{g^{\mathrm{img}}, g^{\mathrm{pt}}, g^{\varnothing}\}$.

To move beyond myopic reactive control, \texttt{WAM-Nav} formulates navigation as conditional asymmetric joint generation. Conditioned on the compact conditioning context $C$ constructed from $(\mathcal{O}_t,\mathcal{S}_t,g)$, the policy predicts a long-horizon action trajectory $\mathbf{A}_t=\{a_t,\ldots,a_{t+H_{\mathrm{act}}-1}\}$ and a short-horizon latent visual foresight sequence $\mathbf{Z}_{t+1:t+H_{\mathrm{vis}}}=\{z_{t+1},\ldots,z_{t+H_{\mathrm{vis}}}\}$, where $H_{\mathrm{vis}}\le H_{\mathrm{act}}$. The learning objective is to model:
\begin{equation}
    p_\theta(\mathbf{A}_t, \mathbf{Z}_{t+1:t+H_{\mathrm{vis}}} \mid C).
\end{equation}
Through this formulation, long-horizon actions maintain trajectory continuity, while short-horizon latent foresight provides reliable near-future geometric constraints for safe navigation.

\begin{algorithm}[b]
\caption{Online Ego-Motion History and Egocentric Conversion}
\begin{algorithmic}
\REQUIRE{window length $k$; executed action $a_{t-1}$; pose accumulator $s_{t-1}$; pose buffer $\mathcal{B}$}
\ENSURE{egocentric relative-motion sequence $\tilde{\mathcal{S}}_t=\{(\Delta x_i,\Delta y_i,\Delta\theta_i)\}_{i=t-k+1}^{t}$}
\STATE $s_t \leftarrow s_{t-1} + a_{t-1}$
\STATE Append $s_t$ to $\mathcal{B}$; if $|\mathcal{B}|>k$, remove the oldest pose
\IF{$|\mathcal{B}|<k$}
    \STATE $\mathcal{S}_t \leftarrow$ zero-pad the front of $\mathcal{B}$ to length $k$
\ELSE
    \STATE $\mathcal{S}_t \leftarrow \mathcal{B}$
\ENDIF
\STATE $s_t=(x_t,y_t,\theta_t) \leftarrow$ the last pose in $\mathcal{S}_t$
\STATE $\tilde{\mathcal{S}}_t \leftarrow \emptyset$
\FOR{each $s_i=(x_i,y_i,\theta_i)$ in $\mathcal{S}_t$}
    \STATE $\delta x \leftarrow x_i - x_t,\;\; \delta y \leftarrow y_i - y_t$
    \STATE $\Delta x \leftarrow \cos\theta_t \cdot \delta x + \sin\theta_t \cdot \delta y$
    \STATE $\Delta y \leftarrow -\sin\theta_t \cdot \delta x + \cos\theta_t \cdot \delta y$
    \STATE $\Delta\theta \leftarrow \mathrm{atan2}(\sin(\theta_i-\theta_t),\,\cos(\theta_i-\theta_t))$
    \STATE Append $(\Delta x,\Delta y,\Delta\theta)$ to $\tilde{\mathcal{S}}_t$
\ENDFOR
\end{algorithmic}
\label{alg:ego_motion}
\end{algorithm}

\section{Method Details}
\label{appendix:method_details}

\subsection{Trajectory-Aware Motion History}
\label{appendix:ego_history_conversion}

While goal-modulated visual memory provides spatial context over $\mathcal{O}_t$, it does not explicitly encode kinematic momentum; the motion trajectory stream therefore maintains an action-accumulated ego-motion history $\mathcal{S}_t=\{s_{t-k+1}, \ldots, s_t\}$ (Sec.~\ref{sec:problem}). At deployment, each executed action $a_{t-1}$ updates the current planar pose $s_t$, which is appended to a length-$k$ sliding buffer and zero-padded at the front when the episode is shorter than $k$ steps. The buffer is then re-expressed in the current egocentric frame---re-centering on $s_t$, rotating translations by $\theta_t$, and wrapping heading differences---so that the stream receives a complete relative-motion sequence $\tilde{\mathcal{S}}_t=\{(\Delta x_i,\Delta y_i,\Delta\theta_i)\}_{i=t-k+1}^{t}$ rather than absolute scene coordinates. Algorithm~\ref{alg:ego_motion} details this online procedure.

\begin{figure}[!b]
    \centering
    \includegraphics[width=0.45\linewidth]{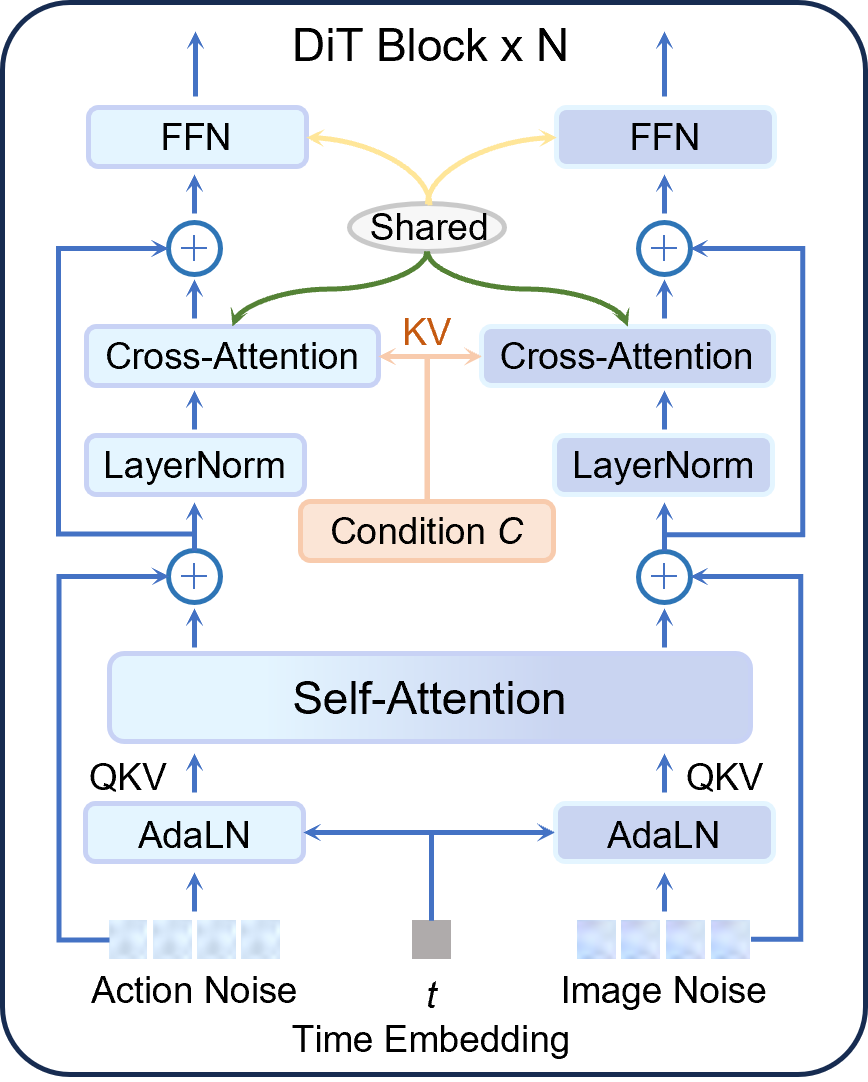}
    \caption{Architecture of the shared DiT block for asymmetric action-foresight generation. Noised action tokens and latent visual foresight tokens are modulated through separate adaLN branches, coupled via shared self-attention, and grounded on the conditioning context $C$ through shared cross-attention and FFN layers.}
    \label{fig:shared_dit}
\end{figure}

\subsection{Architecture of the Shared DiT Block}
\label{appendix:shared_dit}
As introduced in the \textbf{Asymmetric Action-Foresight Generation} subsection, the shared DiT predicts the joint flow-matching velocity fields $\hat{u}_A$ and $\hat{u}_Z$ from the interpolated action and latent visual states $(\mathbf{A}_{\tau}, \mathbf{Z}_{\tau})$, conditioned on the conditioning context $C$ and flow timestep $\tau$. Figure.~\ref{fig:shared_dit} illustrates a single DiT block, stacked $N$ times.

At each block, $\mathbf{A}_{\tau}$ and $\mathbf{Z}_{\tau}$ are tokenized into parallel action and image streams. The timestep embedding $\tau$ is injected into each stream via separate adaLN-Zero branches. The modulated tokens then pass through a shared self-attention layer, enabling early interaction between action and latent visual tokens. After the residual update, both streams perform cross-attention over the conditioning context $C$ using shared cross-attention and FFN weights with stream-specific modulation. The updated features are propagated through stacked blocks, and finally projected to $\hat{u}_A$ and $\hat{u}_Z$. This design enables consistent action--foresight generation through shared parameters and cross-modal coupling.

\section{Implementation Details of WAM-Nav}
\label{appendix:implementation_details}

\begin{table}[t]
\centering
\caption{Key hyperparameters of \texttt{WAM-Nav}.}
\label{tab:hyperparameters}
\vspace{5pt}
\begin{tabular*}{\columnwidth}{@{\extracolsep{\fill}}lc|lc}
\toprule
\textbf{Parameter} & \textbf{Value} & \textbf{Parameter} & \textbf{Value} \\
\midrule
Hidden Dimension & 384 & Condition Length & 64 \\
Memory Window ($k$) & 8 & Action Horizon ($H_{\mathrm{act}}$) & 24 \\
DiT Layers & 16 & DiT Attention Heads & 8 \\
Fusion Layers & 4 & Fusion Attention Heads & 8 \\
Causal Layers & 2 & Causal Attention Heads & 4 \\
Latent Grid Size & $16 \times 16$ & Latent Channels & 4 \\
Image Patches & 16 & Dropout & 0.1 \\
Batch Size & 256 & Learning Rate & $1.5 \times 10^{-4}$ \\
Weight Decay & $1\times10^{-4}$ & Warmup Ratio & 0.05 \\
$\lambda_{\mathrm{img}}$ & 0.25 & $\lambda_{\mathrm{align}}$ & 0.1 \\
Inference Steps & 10 & Visual Horizon ($H_{\mathrm{vis}}$) & 1 \\
Visual Backbone & DINOv2 ViT-S/14 & -- & -- \\
\bottomrule
\end{tabular*}
\end{table}

We train \texttt{WAM-Nav} using a batch size of 256 and optimize the model with the AdamW optimizer. The peak learning rate is set to $1.5 \times 10^{-4}$ with a weight decay of $1 \times 10^{-4}$, following a cosine annealing schedule with a 5\% linear warmup. During training, the visual backbone (DINOv2 ViT-S/14) and the pre-trained Stable Diffusion VAE are kept frozen to preserve their general-purpose representations, while the goal image encoder, fusion decoder, causal motion encoder, and the shared DiT are trained from scratch. The loss weighting coefficients in Eq.~\ref{eq:l_total} are empirically set to $\lambda_{\mathrm{img}}=0.25$ and $\lambda_{\mathrm{align}}=0.1$. At deployment, the flow-matching ODE solver runs for $N_{\mathrm{steps}}=10$ Euler integration steps, which balances high-quality trajectory generation with real-time responsiveness. The detailed hyperparameter configurations are summarized in Table~\ref{tab:hyperparameters}.

\section{Baseline Details}
\label{appendix:baseline_details}
To comprehensively evaluate \texttt{WAM-Nav}, we compare against representative baselines spanning reinforcement learning, reactive planning, foundation models, and generative world models.

\begin{itemize}
    \item \textbf{DD-PPO}~\citep{wijmans2020ddppo}: A large-scale distributed reinforcement learning framework for Point-Goal navigation, trained over billions of frames in Habitat and widely adopted as a strong mapless navigation baseline.

    \item \textbf{iPlanner}~\citep{yang2023iplanner}: A high-frequency local planner that directly predicts multi-step continuous trajectories through spline regression for reactive obstacle avoidance.

    \item \textbf{ViPlanner}~\citep{roth2024viplanner}: An end-to-end semantic planner that predicts 2D collision cost maps from monocular observations for safe trajectory generation.

    \item \textbf{GNM}~\citep{shah2023gnm}: A general navigation foundation model trained on large-scale cross-embodiment datasets for zero-shot transfer and low-level action prediction.

    \item \textbf{ViNT}~\citep{shah2023vint}: A Transformer-based navigation policy that extends GNM by jointly modeling heterogeneous visual navigation data with improved sequence representation.

    \item \textbf{NoMaD}~\citep{sridhar2024nomad}: A diffusion-based action policy that introduces goal masking to unify goal-conditioned navigation and goal-free exploration.

    \item \textbf{NWM}~\citep{bar2025nwm}: A generative navigation world model based on conditional DiT that predicts future egocentric observations for foresight-driven planning.

    \item \textbf{NavDP}~\citep{cai2025navdp}: A diffusion-based navigation policy that leverages privileged simulation priors for zero-shot sim-to-real transfer.
\end{itemize}

\section{Additional Experimental Results}
\label{appendix:additional_results}
\subsection{Main Simulation Results}
\label{appendix:main_results}

\begin{table*}[!h]
\centering
\caption{Performance comparison across three navigation tasks.}
\label{tab:all_tasks_fixed_vertical}

\small
\renewcommand{\arraystretch}{0.8}

\begin{tabular*}{\textwidth}{@{\extracolsep{\fill}}l cc cc cc cc cc}
\toprule

& \multicolumn{4}{c}{ClutterScenes}
& \multicolumn{4}{c}{InternScenes}
& \multicolumn{2}{c}{\multirow{2}{*}{Average}} \\
\cmidrule(lr){2-5}
\cmidrule(lr){6-9}

Method
& \multicolumn{2}{c}{Easy}
& \multicolumn{2}{c}{Hard}
& \multicolumn{2}{c}{Home}
& \multicolumn{2}{c}{Commercial}
& \multicolumn{2}{c}{} \\
\cmidrule(lr){2-3}
\cmidrule(lr){4-5}
\cmidrule(lr){6-7}
\cmidrule(lr){8-9}
\cmidrule(lr){10-11}

& $M_1$ & $M_2$ & $M_1$ & $M_2$ & $M_1$ & $M_2$ & $M_1$ & $M_2$ & $M_1$ & $M_2$ \\

\midrule

\multicolumn{11}{c}{\textbf{Task 1: Image-Goal Navigation} ($M_1$: SR $\uparrow$, $M_2$: SPL $\uparrow$)} \\
\cmidrule(lr){1-11}

GNM & 26.9 & 25.9 & 20.0 & 19.2 & 7.7 & 7.3 & 10.5 & 10.3 & 16.3 & 15.7 \\
ViNT & 15.0 & 13.9 & 16.5 & 15.2 & 6.2 & 6.0 & 12.9 & 12.2 & 12.6 & 11.8 \\
NoMaD & 8.3 & 6.9 & 7.3 & 6.9 & 6.7 & 6.2 & 11.7 & 10.7 & 8.5 & 7.6 \\
NWM & 12.7 & 12.0 & 8.7 & 8.4 & 4.0 & 3.9 & 6.9 & 6.6 & 8.1 & 7.7 \\
NavDP & 49.7 & 48.4 & 47.5 & 46.3 & 29.8 & 27.7 & 46.6 & 43.3 & 43.4 & 41.4 \\
Ours & \textbf{66.1} & \textbf{64.5} & \textbf{55.6} & \textbf{53.4} & \textbf{30.7} & \textbf{29.4} & \textbf{48.3} & \textbf{45.6} & \textbf{50.2} & \textbf{48.2} \\

\midrule

\multicolumn{11}{c}{\textbf{Task 2: Point-Goal Navigation} ($M_1$: SR $\uparrow$, $M_2$: SPL $\uparrow$)} \\
\cmidrule(lr){1-11}

DD-PPO & 0.0 & 0.0 & 0.0 & 0.0 & 0.4 & 0.4 & 5.3 & 5.2 & 1.4 & 1.4 \\
iPlanner & 89.4 & 88.4 & 80.3 & 78.8 & 43.0 & 40.6 & 54.6 & 52.8 & 66.8 & 65.1 \\
ViPlanner & 80.2 & 80.1 & 64.7 & 64.5 & 45.0 & 43.2 & 63.7 & 61.9 & 63.4 & 62.4 \\
NavDP & 92.3 & 90.5 & 87.4 & 84.9 & 60.0 & 55.6 & 71.4 & 68.2 & 77.8 & 74.8 \\
Ours & \textbf{93.8} & \textbf{91.6} & \textbf{88.9} & \textbf{86.4} & \textbf{61.4} & \textbf{58.6} & \textbf{77.4} & \textbf{75.3} & \textbf{80.4} & \textbf{78.0} \\

\midrule

\multicolumn{11}{c}{\textbf{Task 3: No-Goal Exploration} ($M_1$: Area $\uparrow$, $M_2$: Time $\uparrow$)} \\
\cmidrule(lr){1-11}

GNM & 42.8 & 17.1 & 29.6 & 12.5 & 19.8 & 30.8 & 19.2 & 28.1 & 27.8 & 22.1 \\
ViNT & 55.3 & 20.3 & 46.6 & 18.9 & 20.8 & 31.5 & 22.4 & 27.2 & 36.2 & 24.4 \\
NoMaD & 119.4 & 48.1 & 85.7 & 36.6 & 21.7 & 27.8 & 23.4 & 29.7 & 62.5 & 35.5 \\
NavDP & 315.6 & 112.7 & 274.1 & 106.2 & 34.1 & 35.0 & 43.9 & 36.2 & 167.2 & 72.5 \\
Ours & \textbf{320.4} & \textbf{115.2} & \textbf{281.7} & \textbf{108.6} & \textbf{36.8} & \textbf{38.3} & \textbf{45.5} & \textbf{40.2} & \textbf{171.1} & \textbf{75.6} \\

\bottomrule
\end{tabular*}
\end{table*}

\subsection{Main Component Ablation}
\label{appendix:main_ablation}

\begin{table}[!h]
\centering
\caption{Ablation study of different components of \texttt{WAM-Nav} on Image-Goal navigation.}
\label{tab:ablation_study_flattened_onedecimal_final}
\vspace{5pt}
\small
\renewcommand{\arraystretch}{0.85}

\begin{tabular*}{\columnwidth}{@{\extracolsep{\fill}}cc cc cc cc}
\toprule
\multicolumn{2}{c}{Method} &
\multicolumn{2}{c}{ClutterScenes (Avg.)} &
\multicolumn{2}{c}{InternScenes (Avg.)} &
\multicolumn{2}{c}{Overall Average} \\

\cmidrule(lr){1-2}
\cmidrule(lr){3-4}
\cmidrule(lr){5-6}
\cmidrule(lr){7-8}

Motion Traj. & Latent Pred. & SR & SPL & SR & SPL & SR & SPL \\
\midrule

$\times$ & $\times$ & 51.0 & 49.9 & 33.2 & 31.7 & 42.1 & 40.8 \\
$\times$ & $\checkmark$ & 57.4 & 55.5 & 34.0 & 32.2 & 45.7 & 43.8 \\
$\checkmark$ & $\times$ & 57.9 & 57.1 & 30.0 & 28.6 & 43.9 & 42.9 \\

$\checkmark$ & $\checkmark$ &
\textbf{60.9} & \textbf{59.0} &
\textbf{39.5} & \textbf{37.5} &
\textbf{50.2} & \textbf{48.2} \\

\bottomrule
\end{tabular*}
\end{table}

\subsection{Additional Ablation Studies}

To systematically investigate the contribution of each key component in \texttt{WAM-Nav}, we conduct ablation studies on unified goal alignment, dual-stream contextual conditioning, and asymmetric action-foresight generation. 

For unified goal alignment, Table~\ref{tab:ablation_goal_alignment} compares variants that retain only the visual query $g_V$, only the geometric query $g_G$, or both streams. Using only $g_V$ favors Image-Goal navigation, while using only $g_G$ performs best on Point-Goal navigation; however, both single-stream variants become less balanced across task types. Combining $g_V$ and $g_G$ achieves the strongest No-Goal exploration performance and maintains competitive Image-Goal and Point-Goal results, supporting our motivation that visual-semantic and geometric goal cues are complementary for a unified multi-task policy.

To validate the remaining design choices, we conduct ablations on the Image-Goal task, including the historical memory window size ($k$), the DSCC fusion design, the action-foresight coupling architecture, and the visual prediction horizon. As shown in Table~\ref{tab:ablation_combined}, both insufficient and excessive temporal context negatively affect navigation performance: $k=4$ provides limited motion context, while $k=16$ makes the policy less responsive to sudden geometric changes. Removing explicit goal injection (\textit{w/o GI}) also causes a clear performance drop, confirming that goal-aware modulation is necessary for constructing effective conditioning contexts. For action-foresight coupling, the fully shared DiT outperforms decoupled and partially shared variants, indicating that latent foresight is most useful when it directly regularizes action generation through shared representations.

Table~\ref{tab:ablation_horizon_asymmetry} further validates the asymmetric horizon design. Following NavDP, we keep the action horizon fixed at $H_{\mathrm{act}}=24$ to preserve long-horizon trajectory continuity, and vary the visual horizon $H_{\mathrm{vis}}$. The best performance is obtained with short-horizon visual foresight ($H_{\mathrm{vis}}=1$), while longer visual horizons progressively degrade performance. This supports our core design motivation: in navigation, visual foresight should provide reliable near-future geometric constraints rather than long autoregressive visual rollouts, which are more prone to error accumulation under large egocentric viewpoint changes.


\begin{table*}[t]
\centering
\small
\renewcommand{\arraystretch}{1.15}
\caption{Ablation studies on memory window size, DSCC module, and Action-Foresight architecture.}
\label{tab:ablation_combined}

\begin{tabular*}{\textwidth}{@{\extracolsep{\fill}}lcc|lcc|lcc}
\toprule
\multicolumn{3}{c|}{\textbf{Memory Window Size}} &
\multicolumn{3}{c|}{\textbf{DSCC}} &
\multicolumn{3}{c}{\textbf{Action-Foresight}} \\
\cmidrule(r){1-3}
\cmidrule(r){4-6}
\cmidrule(l){7-9}

\textbf{Setting} & SR ($\uparrow$) & SPL ($\uparrow$) &
\textbf{Variant} & SR ($\uparrow$) & SPL ($\uparrow$) &
\textbf{Variant} & SR ($\uparrow$) & SPL ($\uparrow$) \\
\midrule

$k=4$ & 46.6 & 44.1 &
-- & -- & -- &
Decoupled DiT & 45.9 & 44.6 \\

$k=8$ (Ours) & \textbf{50.2} & \textbf{48.2} &
w/o GI & 44.5 & 42.2 &
Partially DiT & 47.5 & 45.4 \\

$k=16$ & 44.9 & 42.3 &
Ours & \textbf{50.2} & \textbf{48.2} &
Ours & \textbf{50.2} & \textbf{48.2} \\

\bottomrule
\end{tabular*}
\end{table*}

\begin{figure}[t]
    \centering
    \includegraphics[width=0.75\linewidth]{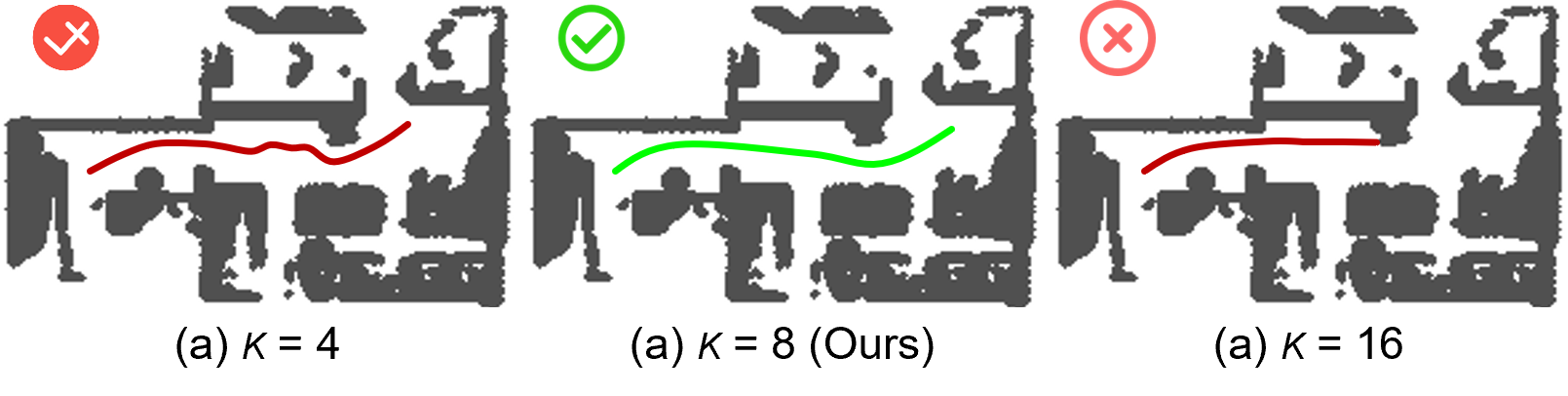}
    \caption{Visualization of the effect of different memory window sizes on navigation trajectory planning.}
    \label{fig:ablation_traj}
\end{figure}

\begin{table}[!t]
\centering
\small
\renewcommand{\arraystretch}{1.05}
\caption{Ablation study of unified goal alignment across three navigation tasks.}
\label{tab:ablation_goal_alignment}
\vspace{5pt}

\begin{tabular*}{\columnwidth}{@{\extracolsep{\fill}}l cc cc cc}
\toprule
\multirow{2}{*}{Variant}
& \multicolumn{2}{c}{Image-Goal}
& \multicolumn{2}{c}{Point-Goal}
& \multicolumn{2}{c}{No-Goal} \\
\cmidrule(lr){2-3}
\cmidrule(lr){4-5}
\cmidrule(lr){6-7}
& SR ($\uparrow$) & SPL ($\uparrow$)
& SR ($\uparrow$) & SPL ($\uparrow$)
& Area ($\uparrow$) & Time ($\uparrow$) \\
\midrule
Only $g_V$ & \textbf{55.6} & \textbf{53.2} & 62.1 & 58.6 & 166.3 & 69.9 \\
Only $g_G$ & 47.9 & 45.3 & \textbf{83.5} & \textbf{80.8} & 168.4 & 72.5 \\
Ours ($g_V+g_G$)
& 50.2 & 48.2
& 80.4 & 78.0
& \textbf{171.1} & \textbf{75.6} \\
\bottomrule
\end{tabular*}
\end{table}

\begin{table}[!t]
\centering
\small
\renewcommand{\arraystretch}{1.05}
\caption{Ablation study of asymmetric action-foresight horizons on Image-Goal navigation.}
\label{tab:ablation_horizon_asymmetry}
\vspace{5pt}

\begin{tabular}{lcc}
\toprule
\textbf{Setting} & SR ($\uparrow$) & SPL ($\uparrow$) \\
\midrule
$H_{\mathrm{act}}=24,\;H_{\mathrm{vis}}=1$ & \textbf{50.2} & \textbf{48.2} \\
$H_{\mathrm{act}}=24,\;H_{\mathrm{vis}}=4$ & 46.2 & 43.8 \\
$H_{\mathrm{act}}=24,\;H_{\mathrm{vis}}=8$ & 39.6 & 37.3 \\
$H_{\mathrm{act}}=24,\;H_{\mathrm{vis}}=24$ & 30.4 & 28.1 \\
\bottomrule
\end{tabular}
\end{table}

\subsection{Comparison with NavDP}
\label{subsec:compared_navdp}

We first evaluate the zero-shot adaptability of a single \texttt{WAM-Nav} policy across diverse robot embodiments without any parameter retraining. As summarized in Table~\ref{tab:embodiment_fixed}, experiments are conducted on a wheeled Dingo platform together with Unitree G1 and H2 humanoid robots in the IsaacSim environment. The quantitative results demonstrate strong cross-embodiment generalization. In particular, on the G1 platform, \texttt{WAM-Nav} consistently outperforms NavDP across all evaluated task settings, achieving a $42.9\%$ SR on Image-Goal navigation. On the more challenging H2 humanoid platform, \texttt{WAM-Nav} also achieves higher performance in all navigation tasks. These results suggest that \texttt{WAM-Nav} effectively learns embodiment-agnostic navigation representations, enabling stable policy transfer across robots with substantially different kinematic structures without retraining.

To further characterize the capabilities of our framework, we compare \texttt{WAM-Nav} and NavDP across navigation tasks of varying difficulty under the Image-Goal setting. As shown in Table~\ref{tab:difficulty}, navigation episodes are divided into Easy, Medium, and Hard subsets according to the ground-truth path length. Across all difficulty levels, \texttt{WAM-Nav} consistently outperforms NavDP, with the performance gap increasing as task difficulty grows. Notably, in the Hard subset, \texttt{WAM-Nav} reaches $25.0\%$ SR and $23.8\%$ SPL, surpassing NavDP by $7.6\%$ and $2.5\%$, respectively.

This growing advantage under long-horizon navigation highlights the benefit of our asymmetric action-foresight generation framework. Over extended trajectories, reactive diffusion policies such as NavDP tend to accumulate execution drift and are more vulnerable to local geometric traps due to the lack of predictive perceptual feedback. In contrast, \texttt{WAM-Nav} jointly models action generation and latent visual foresight, allowing near-future visual predictions to regularize trajectory planning. This coupled predictive mechanism enforces long-range spatiotemporal consistency, reduces myopic behavioral drift, and enables the policy to maintain stable and goal-directed navigation even in long-distance and cluttered environments.

\begin{table}[t]
\centering
\caption{Generalization performance across different embodiments.}
\label{tab:embodiment_fixed}
\vspace{5pt}

\small
\renewcommand{\arraystretch}{1.05}

\begin{tabular*}{\columnwidth}{@{\extracolsep{\fill}}cl cc cc cc}
\toprule

\multirow{2}{*}{Embodiment}
& \multirow{2}{*}{Method}
& \multicolumn{2}{c}{Image-Goal}
& \multicolumn{2}{c}{Point-Goal}
& \multicolumn{2}{c}{No-Goal} \\

\cmidrule(lr){3-4}
\cmidrule(lr){5-6}
\cmidrule(lr){7-8}

&
& SR ($\uparrow$) & SPL ($\uparrow$)
& SR ($\uparrow$) & SPL ($\uparrow$)
& Area ($\uparrow$) & Time ($\uparrow$) \\

\midrule

\multirow{2}{*}{Dingo}
& NavDP
& 43.4 & 41.4
& 77.8 & 74.8
& 167.2 & 72.5 \\

& Ours
& \textbf{50.2} & \textbf{48.2}
& \textbf{80.4} & \textbf{78.0}
& \textbf{171.1} & \textbf{75.6} \\

\midrule

\multirow{2}{*}{G1}
& NavDP
& 37.2 & 35.6
& 60.3 & 56.8
& 133.0 & 65.2 \\

& Ours
& \textbf{42.9} & \textbf{42.5}
& \textbf{64.8} & \textbf{63.1}
& \textbf{147.6} & \textbf{68.3} \\

\midrule

\multirow{2}{*}{H2}
& NavDP
& 32.7 & 31.3
& 47.9 & 46.5
& 142.4 & 67.6 \\

& Ours
& \textbf{35.5} & \textbf{34.9}
& \textbf{52.7} & \textbf{50.5}
& \textbf{145.4} & \textbf{70.6} \\

\bottomrule
\end{tabular*}

\end{table}

\begin{table}[t]
\centering
\caption{Performance across navigation difficulty levels.}
\label{tab:difficulty}
\vspace{5pt}

\small
\setlength{\tabcolsep}{7pt}
\renewcommand{\arraystretch}{1.05}

\begin{tabular*}{\columnwidth}{@{\extracolsep{\fill}}l cc cc cc cc}
\toprule
\multirow{2}{*}{Method}
& \multicolumn{2}{c}{Easy}
& \multicolumn{2}{c}{Medium}
& \multicolumn{2}{c}{Hard}
& \multicolumn{2}{c}{Average} \\
\cmidrule(lr){2-3}
\cmidrule(lr){4-5}
\cmidrule(lr){6-7}
\cmidrule(lr){8-9}
& SR ($\uparrow$) & SPL ($\uparrow$)
& SR ($\uparrow$) & SPL ($\uparrow$)
& SR ($\uparrow$) & SPL ($\uparrow$)
& SR ($\uparrow$) & SPL ($\uparrow$) \\
\midrule
NavDP & 46.2 & 45.7 & 33.8 & 35.6 & 17.4 & 21.3 & 43.4 & 41.4 \\
Ours  & \textbf{52.5} & \textbf{50.5}
      & \textbf{42.4} & \textbf{40.5}
      & \textbf{25.0} & \textbf{23.8}
      & \textbf{50.2} & \textbf{48.2} \\
\bottomrule
\end{tabular*}
\end{table}

\section{Real-World Deployment Details}
\label{appendix:realworld_details}
\subsection{Hardware setup}
\begin{figure}[t]
    \centering
    \includegraphics[width=0.75\linewidth]{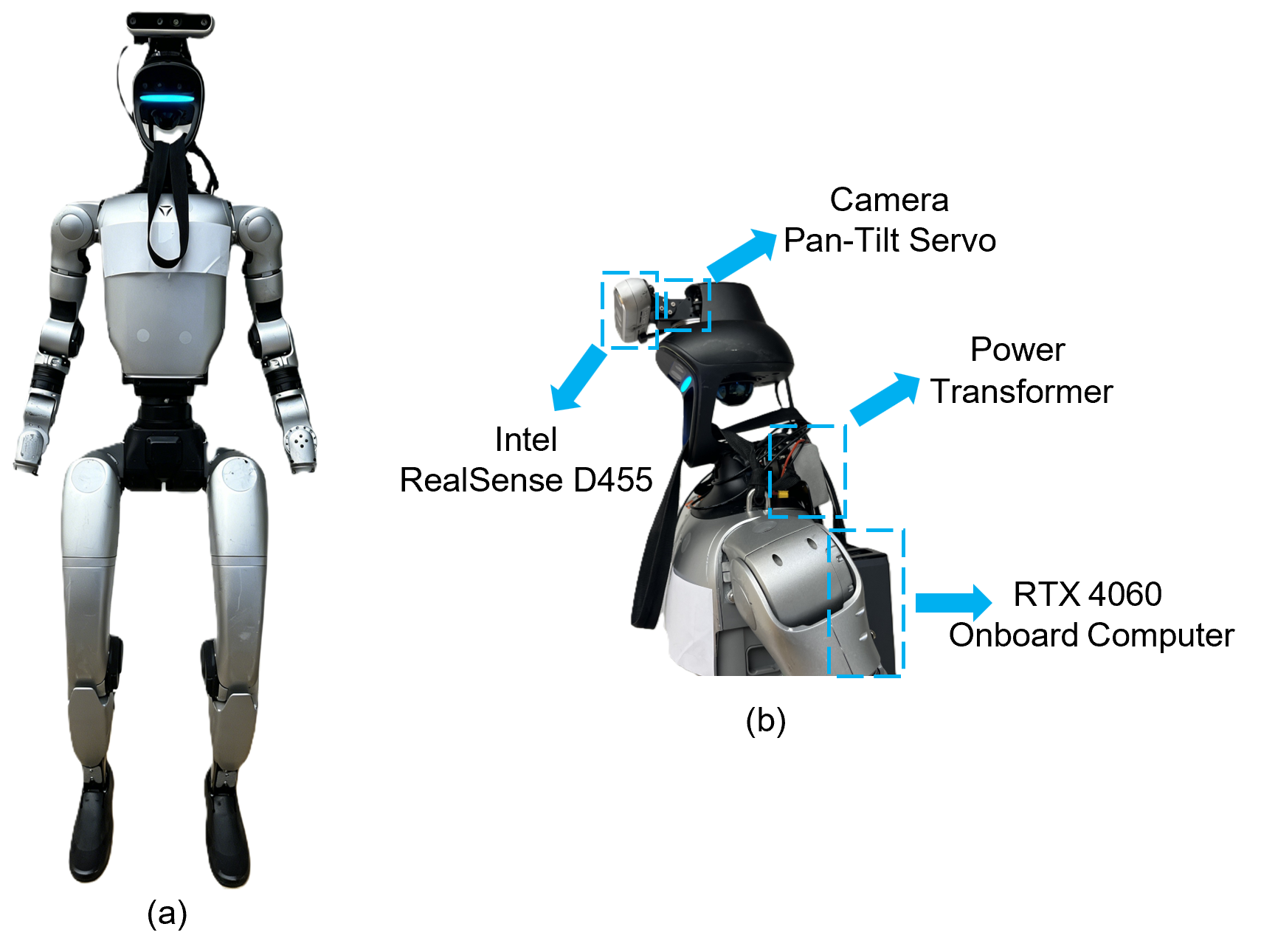}
    \caption{Robot Setup. (a) Unitree G1. (b) Detailed view of the hardware components.}
    \label{fig:G1}
\end{figure}

As shown in Figure.~\ref{fig:G1}, our experimental platform is built on the Unitree G1 humanoid robot. The robot is equipped with an Intel RealSense D455 depth camera mounted on a pan-tilt servo to capture RGB-D observations from fixed viewpoints. A backpack-mounted onboard computer equipped with an RTX 4060 GPU performs real-time perception and control. A power transformer provides stable power supply for the onboard devices. This hardware setup enables real-time perception, computation, and execution for our experiments.

\subsection{Deployment Details}
During deployment, the Intel RealSense D455 depth camera is mounted on the pan-tilt servo and configured with a 20° downward pitch angle to improve the observation of ground-level obstacles. The camera captures RGB-D observations at 30 Hz.
The WAM-Nav model is deployed as a server on the backpack-mounted onboard computer equipped with an RTX 4060 GPU for online inference. The model occupies approximately 1.3 GB of GPU memory, and the inference frequency is set to 1 Hz.
The client module is also deployed on the onboard computer. It is responsible for forwarding the incoming observations to the server and receiving the predicted outputs. The predicted trajectories are tracked using Model Predictive Control (MPC), which further converts them into velocity commands for the robot motors. During experiments, the control loop runs at 50 Hz, with each command applied for 0.1 s.

\section{Additional Failure Case Analysis}
\label{appendix:failure_analysis}
\begin{figure}[!h]
    \centering
    \includegraphics[width=0.75\linewidth]{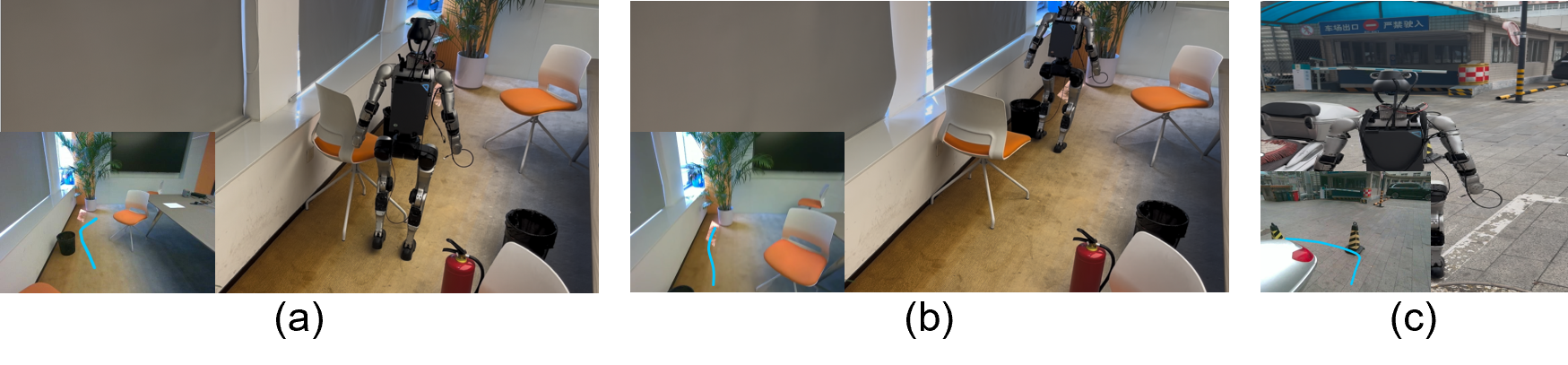}
    \caption{Failure Cases. The bottom-left subfigure shows the first-person view of the planned trajectory}
    \label{fig:failure}
\end{figure}
In Figure~\ref{fig:failure}, we visualize several typical failure cases. Case (a) shows that when the robot is close to a low obstacle, the limited field of view results in a weak perception of the obstacle, causing the robot to overlook it and eventually collide. Another common type of failure is illustrated in (b) and (c): when obstacles are located on the left or right side of the robot, the robot’s body shape is not taken into account. As a result, trajectory planning only ensures that the camera can pass through, leading to collisions between the robot and the obstacles.
\end{document}